\documentclass[conference]{IEEEtran}
\IEEEoverridecommandlockouts
\usepackage{cite}
\usepackage{amsmath,amssymb,amsfonts}
\usepackage{algorithmic}
\usepackage{filecontents}
\usepackage{graphicx}
\usepackage{textcomp}
\usepackage{xcolor}
\usepackage{booktabs} 
\usepackage{multirow}
\usepackage{cleveref}

\def\BibTeX{{\rm B\kern-.05em{\sc i\kern-.025em b}\kern-.08em
    T\kern-.1667em\lower.7ex\hbox{E}\kern-.125emX}}

\begin{document}

\title{Emotion helps Sentiment: A Multi-task Model for Sentiment and Emotion Analysis}

\author{\IEEEauthorblockN{Abhishek Kumar}
\IEEEauthorblockA{\textit{Department of Computer Science and Engineering} \\
\textit{Indian Institute of Technology Patna, India}\\
abhishek.ee14@iitp.ac.in}
\and
\IEEEauthorblockN{Asif Ekbal}
\IEEEauthorblockA{\textit{Department of Computer Science and Engineering} \\
\textit{Indian Institute of Technology Patna, India}\\
asif@iitp.ac.in}
\and
\IEEEauthorblockN{Daisuke Kawahra}
\IEEEauthorblockA{\textit{Department of Intelligence Science and Technology } \\
\textit{Kyoto University, Japan}\\
dk@i.kyoto-u.ac.jp}
\and
\IEEEauthorblockN{Sadao Kurohashi}
\IEEEauthorblockA{\textit{Department of Intelligence Science and Technology } \\
\textit{Kyoto University, Japan}\\
kuro@i.kyoto-u.ac.jp}
}

\maketitle

\begin{abstract}
In this paper, we propose a two-layered multi-task attention based neural network that performs sentiment analysis through emotion analysis. The proposed approach is based on Bidirectional Long Short-Term Memory and uses Distributional Thesaurus as a source of external knowledge to improve the sentiment and emotion prediction. The proposed system has two levels of attention to hierarchically build a meaningful representation. We evaluate our system on the benchmark dataset of SemEval 2016 Task 6 and also compare it with the state-of-the-art systems on Stance Sentiment Emotion Corpus. Experimental results show that the proposed system improves the performance of sentiment analysis by 3.2 F-score points on SemEval 2016 Task 6 dataset. Our network also boosts the performance of emotion analysis by 5 F-score points on Stance Sentiment Emotion Corpus. 
\end{abstract}


\section{Introduction}
The emergence of social media sites with limited character constraint has ushered in a new style of communication. Twitter users within 280 characters per tweet share meaningful and informative messages. These short messages have a powerful impact on how we perceive and interact with other human beings. Their compact nature allows them to be transmitted efficiently and assimilated easily. These short messages can shape people's thought and opinion. This makes them an interesting and important area of study. Tweets are not only important for an individual but also for the companies, political parties or any organization. Companies can use tweets to gauge the performance of their products and predict market trends \cite{goonatilake2007volatility}. The public opinion is particularly interesting for political parties as it gives them an idea of voter's inclination and their support. Sentiment and emotion analysis can help to gauge product perception, predict stock prices and model public opinions \cite{si2013exploiting}.

Sentiment analysis \cite{pang2008opinion} is an important area of research in natural language processing (NLP) where we automatically determine the sentiments (\textit{positive}, \textit{negative}, \textit{neutral}). Emotion analysis focuses on the extraction of predefined emotion from documents. Discrete emotions \cite{dalgleish2000handbook,plutchik2001nature} are often classified into \textit{anger}, \textit{anticipation}, \textit{disgust}, \textit{fear}, \textit{joy}, \textit{sadness}, \textit{surprise} and \textit{trust}. Sentiments and emotions are subjective and hence they are understood similarly and often used interchangeably. This is also mostly because both emotions and sentiments refer to experiences that result from the combined influences of the biological, the cognitive, and the social \cite{delamater2006handbook}. However, emotions are brief episodes and are shorter in length \cite{munezero2014they}, whereas sentiments are formed and retained for a longer period. Moreover, emotions are not always target-centric whereas sentiments are directed. Another difference between emotion and sentiment is that a sentence or a document may contain multiple emotions but a single overall sentiment.

Prior studies show that sentiment and emotion are generally tackled as two separate problems. Although sentiment and emotion are not exactly the same, they are closely related. Emotions, like \textit{joy} and \textit{trust}, intrinsically have an association with a \textit{positive} sentiment. Similarly, \textit{anger}, \textit{disgust}, \textit{fear} and \textit{sadness} have a \textit{negative} tone. Moreover, sentiment analysis alone is insufficient at times in imparting complete information. A \textit{negative} sentiment can arise due to \textit{anger}, \textit{disgust}, \textit{fear}, \textit{sadness} or a combination of these. Information about emotion along with sentiment helps to better understand the state of the person or object. The close association of emotion with sentiment motivates us to build a system for sentiment analysis using the information obtained from emotion analysis. 

In this paper, we put forward a robust two-layered multi-task attention based neural network which performs sentiment analysis and emotion analysis simultaneously. 
The model uses two levels of attention - the first primary attention builds the best representation for each word using Distributional Thesaurus and the secondary attention mechanism creates the final sentence level representation. The system builds the representation hierarchically which gives it a good intuitive working insight. We perform several experiments to evaluate the usefulness of primary attention mechanism. Experimental results show that the two-layered multi-task system for sentiment analysis which uses emotion analysis as an auxiliary task improves over the existing state-of-the-art system of SemEval 2016 Task 6 \cite{mohammad2017stance}.

The main contributions of the current work are two-fold: \textbf{a)} We propose a novel two-layered multi-task attention based system for joint sentiment and emotion analysis. This system has two levels of attention which builds a hierarchical representation. This provides an intuitive explanation of its working; \textbf{b)} We empirically show that emotion analysis is relevant and useful in sentiment analysis. The multi-task system utilizing fine-grained information of emotion analysis performs better than the single task system of sentiment analysis.

\section{Related Work}

A survey of related literature reveals the use of both classical and deep-learning approaches for sentiment and emotion analysis. The system proposed in \cite{kiritchenko2014sentiment} relied on supervised statistical text classification which leveraged a variety of surface form,
semantic, and sentiment features for short informal texts. A Support Vector Machine (SVM) based system for sentiment analysis was used in \cite{kumar2017iitpb}, whereas an ensemble of four different sub-systems for sentiment analysis was proposed in \cite{akhtar2017multilayer}. It comprised of Long Short-Term Memory (LSTM) \cite{Hochreiter:1997:LSM:1246443.1246450}, Gated Recurrent Unit (GRU) \cite{Cho-GRU}, Convolutional Neural Network (CNN) \cite{cnn2014} and Support Vector Regression (SVR) \cite{svr:2004}. \cite{schuff2017annotation} reported the results for emotion analysis using SVR, LSTM, CNN and Bi-directional LSTM (Bi-LSTM) \cite{Graves:2005:BLN:1986079.1986220}. \cite{bandhakavi2017lexicon} proposed a lexicon based feature extraction for emotion text classification. A rule-based approach was adopted by \cite{Wu:2006:ERT:1165255.1165259} to extract emotion-specific semantics. \cite{ho2012high} used a high-order Hidden Markov Model (HMM) for emotion detection. \cite{kahou2016emotion} explored deep learning techniques for end-to-end trainable emotion recognition. \cite{Balikas:2017:MLF:3077136.3080702} proposed a multi-task learning model for fine-grained sentiment analysis. They used ternary sentiment classification \textit{(negative, neutral, positive)} as an auxiliary task for fine-grained sentiment analysis \textit{(very-negative, negative, neutral, positive, very-positive)}. A CNN based system was proposed by \cite{deriu2016sentiment} for three phase joint multi-task training. \cite{You:2016:SEA:2964284.2971475} presented a multi-task learning based model for joint sentiment analysis and semantic embedding learning tasks. \cite{beck2014joint} proposed a multi-task setting for emotion analysis based on a vector-valued Gaussian Process (GP) approach known as coregionalisation \cite{alvarez2012kernels}. A hierarchical document classification system based on sentence and document representation was proposed by \cite{yang2016hierarchical}. An attention framework for sentiment regression is described in \cite{kumar2018knowledge}.
\cite{felbo2017using} proposed a DeepEmoji system based on transfer learning for sentiment, emotion and sarcasm detection through emoji prediction. However, the DeepEmoji system treats these independently, one at a time. 

Our proposed system differs from the above works in the sense that none of these works addresses the problem of sentiment and emotion analysis concurrently. Our empirical analysis shows that performance of sentiment analysis is boosted significantly when this is jointly performed with emotion analysis. This may be because of the fine-grained characteristics of emotion analysis that provides useful evidences for sentiment analysis. 

\section{Proposed Methodology}
\begin{figure*}[h!]
\begin{center}
\includegraphics[width=0.58\textwidth]{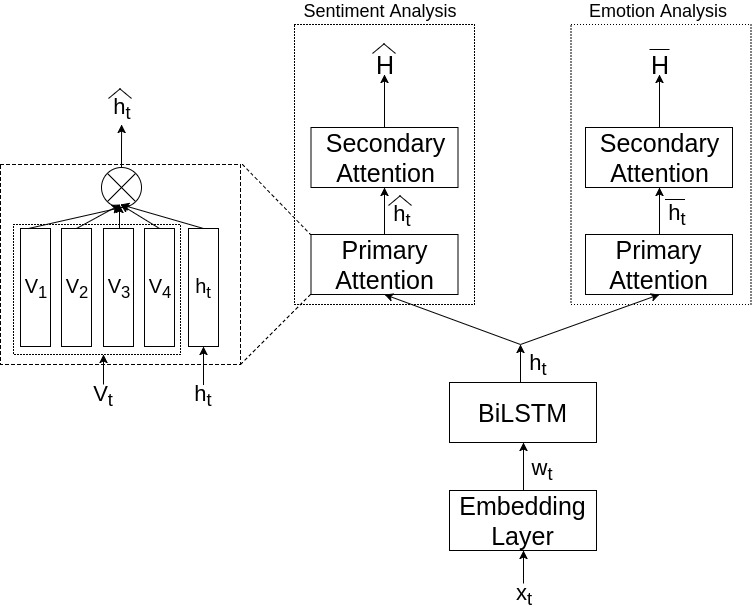}
\caption{Two-layered multi-task attention based network}
\label{fig:model}
\end{center}
\end{figure*}

We propose a novel two-layered multi-task attention based neural network for sentiment analysis where emotion analysis is utilized to improve its efficiency. Figure \ref{fig:model} illustrates the overall architecture of the proposed multi-task system. 
The proposed system consists of a Bi-directional Long Short-Term Memory (BiLSTM) \cite{Graves:2005:BLN:1986079.1986220}, a two-level attention mechanism \cite{bahdanau2014neural,luong2015effective} and a shared representation for emotion and sentiment analysis tasks. The BiLSTM encodes the word representation of each word. This representation is shared between the subsystems of sentiment and emotion analysis. Each of the shared representations is then fed to the primary attention mechanism of both the subsystems. The primary attention mechanism finds the best representation for each word for each task. The secondary attention mechanism acts on top of the primary attention to extract the best sentence representation by focusing on the suitable context for each task. Finally, the representations of both the tasks are fed to two different feed-forward neural networks to produce two outputs - one for sentiment analysis and one for emotion analysis. Each component is explained in the subsequent subsections.

\subsection {Two-Layered Multi-Task Attention Model}
\subsubsection{BiLSTM based word encoder}
Recurrent Neural Networks (RNN) are a class of networks which take sequential input and computes a hidden state vector for each time step. The current hidden state vector depends on the current input and the previous hidden state vector. This makes them good for handling sequential data. However, they suffer from a vanishing or exploding gradient problem when presented with long sequences. The gradient for back-propagating error either reduces to a very small number or increases to a very high value which hinders the learning process. Long Short Term Memory (LSTM) \cite{Hochreiter:1997:LSM:1246443.1246450}, a variant of RNN solves this problem by the gating mechanisms. The input, forget and output gates control the information flow.

BiLSTM is a special type of LSTM which takes into account the output of two LSTMs - one working in the forward direction and one working in the backward direction. The presence of contextual information for both past and future helps the BiLSTM to make an informed decision. The concatenation of a hidden state vectors $\overrightarrow{h_t}$ of the forward LSTM and $\overleftarrow{h_t}$ of the backward LSTM at any time step \textit{t} provides the complete information. Therefore, the output of the BiLSTM at any time step \textit{t} is $h_t$ = [$\overrightarrow{h_t}$, $\overleftarrow{h_t}$]. The output of the BiLSTM is shared between the main task (Sentiment Analysis) and the auxiliary task (Emotion Analysis).

\subsubsection{Word Attention}

The word level attention (primary attention) mechanism gives the model a flexibility to represent each word for each task differently. This improves the word representation as the model chooses the best representation for each word for each task. A Distributional Thesaurus (DT) identifies words that are semantically similar, based on whether they tend to occur in a similar context. It provides a word expansion list for words based on their contextual similarity. We use the top-4 words for each word as their candidate terms. We only use the top-4 words for each word as we observed that the expansion list with more words started to contain the antonyms of the current word which empirically reduced the system performance. Word embeddings of these four candidate terms and the hidden state vector $h_t$ of the input word are fed to the primary attention mechanism. The primary attention mechanism finds the best attention coefficient for each candidate term. At each time step $t$ we get V($x_t$) candidate terms for each input $x_t$ with $v_i$ being the embedding for each term (Distributional Thesaurus and word embeddings are described in the next section). The primary attention mechanism assigns an attention coefficient to each of the candidate terms having the index $i$ $\in$ V($x_t$):

\begin{equation}
\alpha_{ti} \propto \exp((h_{t}^TW_{w}+b_{w})v_i) \label{eq:1}
\end{equation}

\noindent where $W_w$ and $b_{w}$ are jointly learned parameters.

\begin{equation}
m_t = \sum_{i\in{V(x_t)}}\alpha_{ti}v_i\label{eq:2}
\end{equation}

\noindent Each embedding of the candidate term is weighted with the attention score $\alpha_{ti}$ and then summed up. This produces $m_{t}$, the representation for the current input $x_{t}$ obtained from the Distributional Thesaurus using the candidate terms.

\begin{equation}
\widehat{h_t} = m_t+h_{t}\label{eq:3}
\end{equation}

\noindent Finally, $m_{t}$ and $h_{t}$ are concatenated to get $\widehat{h_{t}}$, the final output of the primary attention mechanism. 

\subsubsection{Sentence Attention}

The sentence attention (secondary attention) part focuses on each word of the sentence and assigns the attention coefficients. The attention coefficients are assigned on the basis of words' importance and their contextual relevance. This helps the model to build the overall sentence representation by capturing the context while weighing different word representations individually. The final sentence representation is obtained by multiplying each word vector representation with their attention coefficient and summing them over. The attention coefficient $\alpha_t$ for each word vector representation and the sentence representation $\widehat{H}$ are calculated as:

\begin{equation}
\alpha_{t} \propto \exp(\tanh(\widehat{h_{t}^T}W_{s}+b_{s})) \label{eq:4}
\end{equation}

\noindent where $W_s$ and $b_{s}$ are parameters to be learned.

\begin{equation}
\widehat{H} = \sum_{t}\alpha_{t}\widehat{h_{t}}\label{eq:5}
\end{equation}

\noindent $\widehat{H}$ denotes the sentence representation for sentiment analysis. Similarly, we calculate $\bar{H}$ which represents the sentence for emotion classification. The system has the flexibility to compute different representations for sentiment and emotion analysis both.

\subsubsection{Final Output}

The final outputs for both sentiment and emotion analysis are computed by feeding $\widehat{H}$ and $\bar{H}$ to two different one-layer feed forward neural networks. For our task, the feed forward network for sentiment analysis has two output units, whereas the feed forward network for emotion analysis has eight output nodes performing multi-label classification.

\subsection{Distributional Thesaurus}
Distributional Thesaurus (DT) \cite{DBLP:journals/jlm/BiemannR13} ranks words according to their semantic similarity. It is a resource which produces a list of words in decreasing order of their similarity for each word. We use the DT to expand each word of the sentence. The top-4 words serve as the candidate terms for each word. For example, the candidate terms for the word \textit{good} are: \textit{great}, \textit{nice} \textit{awesome} and \textit{superb}. DT offers the primary attention mechanism external knowledge in the form of candidate terms. It assists the system to perform better when presented with unseen words during testing as the unseen words could have been a part of the DT expansion list. For example, the system may not come across the word \textit{superb} during training but it can appear in the test set. Since the system has already seen the word \textit{superb} in the DT expansion list of the word \textit{good}, it can handle this case efficiently. This fact is established by our evaluation results as the model performs better when the DT expansion and primary attentions are a part of the final multi-task system.

\subsection{Word Embeddings}
Word embeddings represent words in a low-dimensional numerical form. They are useful for solving many NLP problems. We use the pre-trained 300 dimensional Google Word2Vec \cite{mikolov2013distributed} embeddings. The word embedding for each word in the sentence is fed to the BiLSTM network to get the current hidden state. Moreover, the primary attention mechanism is also applied to the word embeddings of the candidate terms for the current word. 

\section{Datasets, Experiments and Analysis}
In this section we present the details of the datasets used for the experiments, results that we obtain and the necessary analysis.
\subsection{Datasets}

We evaluate our proposed approach for joint sentiment and emotion analysis on the benchmark dataset of SemEval 2016 Task 6 \cite{mohammad2017stance} and Stance Sentiment Emotion Corpus (SSEC) \cite{schuff2017annotation}. The SSEC corpus is an annotation of the SemEval 2016 Task 6 corpus with emotion labels. The re-annotation of the SemEval 2016 Task 6 corpus helps to bridge the gap between the unavailability of a corpus with sentiment and emotion labels. The SemEval 2016 corpus contains tweets which are classified into \textit{positive}, \textit{negative} or \textit{other}. It contains 2,914 training and 1,956 test instances. The SSEC corpus is annotated with  \textit{anger}, \textit{anticipation}, \textit{disgust}, \textit{fear}, \textit{joy}, \textit{sadness}, \textit{surprise} and \textit{trust} labels. Each tweet could belong to one or more emotion classes and one sentiment class. Table \ref{data} shows the data statistics of SemEval 2016 task 6 and SSEC which are used for sentiment and emotion analysis, respectively.

\begin{table*}[h]
\begin{center}
\resizebox{0.89\textwidth}{!}{
\begin{tabular}{|l|c|c|c|c|c|c|c|c|c|c|c|} \hline 
\multirow{3}{*}{Data} & \multicolumn{3}{|c|}{Sentiment Dataset} & \multicolumn{8}{|c|}{Emotion Dataset} \\
 & \multicolumn{3}{|c|}{(SemEval 2016 task 6)} & \multicolumn{8}{|c|}{(Stance Sentiment Emotion Corpus)} \\ \cline{2-12}
 & pos & neg & other & anger & anticipation & disgust & fear & joy & sadness & surprise & trust \\ \hline \hline
Train & 963 & 1762 & 189 & 1657 & 1495 & 1271 & 1040 & 1310 & 1583 & 581 & 1032 \\
Test  & 561 & 1272 & 123 & 1245 & 1205 & 912 & 800 & 757 & 1061 & 527 & 681 \\ \hline
\end{tabular}}
\end{center}
\caption{Dataset statistics of SemEval 2016 Task 6 and SSEC used for sentiment and emotion analysis, respectively.}
\label{data}
\end{table*}

\subsection{Preprocessing}

The SemEval 2016 task 6 corpus contains tweets from Twitter. Since the tweets are derived from an environment with the constraint on the number of characters, there is an inherent problem of word concatenation, contractions and use of hashtags. Example: \#BeautifulDay, we've, etc. Usernames and URLs do not impart any sentiment and emotion information (e.g. @John). We use the Python package \textit{ekphrasis} \cite{baziotis-pelekis-doulkeridis:2017:SemEval2} for handling these situations. Ekphrasis helps to split the concatenated words into individual words and expand the contractions. For example, \textit{\#BeautifulDay} to \textit{\# Beautiful Day} and \textit{we've} to \textit{we have}. We replace usernames with \textit{$<$user$>$}, number with \textit{$<number>$} and URLs with \textit{$<$url$>$} token.

\subsection{Implementation Details}
We implement our model in Python using Tensorflow on a single GPU. We experiment with six different BiLSTM based architectures. The three architectures correspond to BiLSTM based systems without primary attention i.e. only with secondary attention for sentiment analysis (S1), emotion analysis (E1) and the multi-task system (M1) for joint sentiment and emotion analysis. The remaining three architectures correspond to the systems for sentiment analysis (S2), emotion analysis (E2) and multi-task system (M2), with both primary and secondary attention. The weight matrices were initialized randomly using numbers form a truncated normal distribution. The batch size was 64 and the dropout \cite{dropout} was 0.6 with the Adam optimizer \cite{adam}. The hidden state vectors of both the forward and backward LSTM were 300-dimensional, whereas the context vector was 150-dimensional. Relu \cite{glorot2011deep} was used as the activation for the hidden layers, whereas in the output layer we used sigmoid as the activation function. Sigmoid cross-entropy was used as the loss function. F1-score was reported for the sentiment analysis \cite{mohammad2017stance} and precision, recall and F1-score were used as the evaluation metric for emotion analysis \cite{schuff2017annotation}. Therefore, we report the F1-score for sentiment and precision, recall and F1-score for emotion analysis.

\begin{table}[h]
\begin{center}
\resizebox{0.47\textwidth}{!}{
\begin{tabular}{|l|l|c|c|}
\hline 
& \bf Models & \bf Sentiment & \bf Emotion \\ \hline \hline
\multicolumn{4}{|l|}{Single task system for Sentiment Analysis} \\ \hline
S1 & only secondary attention & 75.37 & - \\
S2 & primary + secondary attention &  77.58 &  - \\ \hline
\multicolumn{4}{|l|}{Single task system for Emotion Analysis} \\ \hline
E1 & only secondary attention & - & 64.94 \\
E2 & primary + secondary attention &  - &  66.66 \\ \hline
\multicolumn{4}{|l|}{Multi-task system} \\ \hline
M1 & only secondary attention & 81.17 & 63.02 \\ 
M2 & primary + secondary attention & 82.10 & 65.44 \\ \hline
\end{tabular}}
\end{center}
\caption{F-score of various models on sentiment and emotion test dataset.}
\label{result-test}
\end{table}

\begin{table}[h]
\begin{center}
\resizebox{0.47\textwidth}{!}{
\begin{tabular}{|l|c|}
\hline 
\bf Models & \textbf{Sentiment} (F-score) \\ \hline \hline
UWB \cite{krejzl2016uwb} & 42.02 \\
INF-UFRGS-OPINION-MINING \cite{dias2016inf} & 42.32 \\
LitisMind & 44.66 \\ 
pkudblab \cite{wei2016pkudblab} &  56.28 \\ 
SVM + n-grams + sentiment \cite{mohammad2017stance} & 78.90 \\
M2 (proposed) & \textbf{82.10} \\ \hline
\end{tabular}}
\end{center}
\caption{Comparison with the state-of-the-art systems of SemEval 2016 task 6 on sentiment dataset.}
\label{result-sentiment-test}
\end{table}

\subsection{Results and Analysis}
We compare the performance of our proposed system with the state-of-the-art systems of SemEval 2016 Task 6 and the systems of \cite{schuff2017annotation}. Experimental results show that the proposed system improves the existing state-of-the-art systems for sentiment and emotion analysis. We summarize the results of evaluation in Table \ref{result-test}. 

\begin{figure}[h!]
\centering
\includegraphics[width=0.48\textwidth]{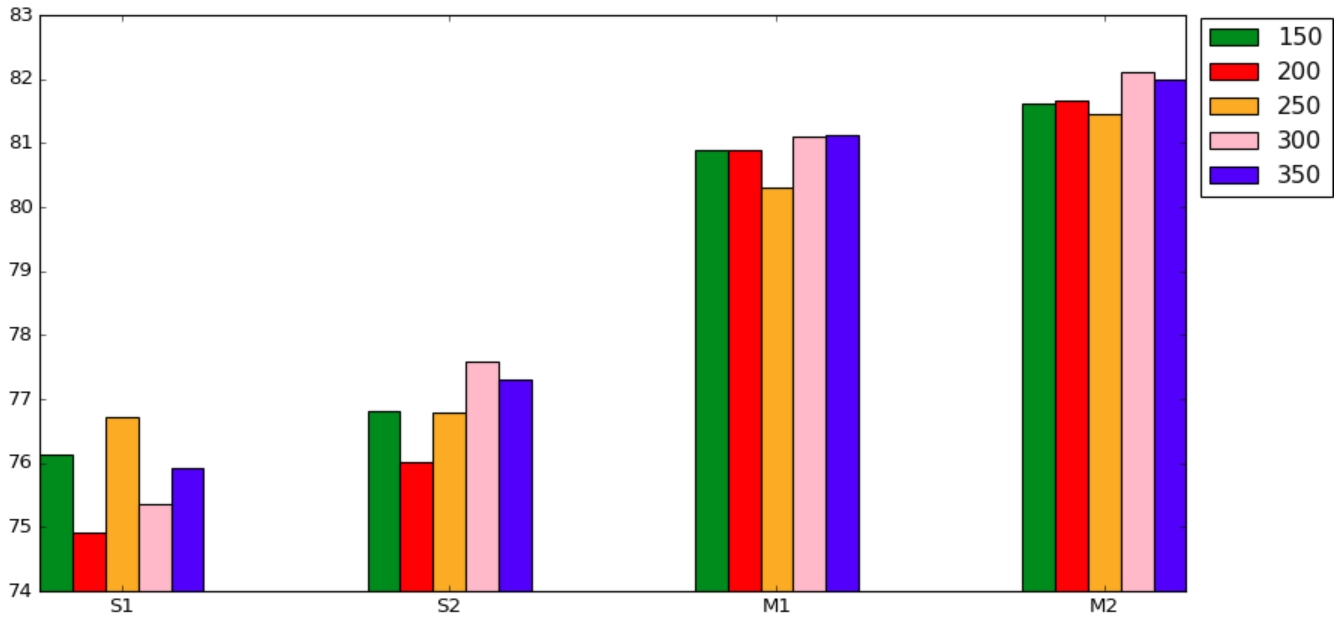}
\caption{Comparison of various models (S1, S2, M1, M2) \textit{w.r.t} different hidden state vector sizes of BiLSTM for sentiment analysis. Y-axis denotes the F-scores.}
\label{sentiment-fig}
\end{figure}

The primary attention mechanism plays a key role in the overall system as it improves the score of both sentiment and emotion analysis in both single task as well as multi-task systems. The use of primary attention improves the performance of single task systems for sentiment and emotion analysis 
by 2.21 and 1.72 points, respectively.
Similarly, when sentiment and emotion analysis are jointly performed the primary attention mechanism improves the score by 0.93 and 2.42 points for sentiment and emotion task, respectively. To further measure the usefulness of the primary attention mechanism and the Distributional Thesaurus, we remove it from the systems S2, E2, and M2 to get the systems S1, E1, and M1. In all the cases, with the removal of primary attention mechanism, the performance drops. This is clearly illustrated in Figure \ref{sentiment-fig}. These observations indicate that the primary attention mechanism is an important component of the two-layered multi-task attention based network for sentiment analysis. We also perform t-test \cite{church1991using} for computing statistical significance of the obtained results from the final two-layered multi-task system M2 for sentiment analysis by calculating the \textit{p-values} and observe that the performance gain over M1 is significant with p-value = 0.001495. Similarly, we perform the statistical significance test for each emotion class. The p-values for anger, anticipation, fear, disgust, joy, sadness, surprise and trust are 0.000002, 0.000143, 0.00403, 0.000015, 0.004607, 0.069, 0.000001 and 0.000001, respectively. These results provide a good indication of statistical significance.

Table \ref{result-sentiment-test} shows the comparison of our proposed system with the existing state-of-the-art system of SemEval 2016 Task 6 for the sentiment dataset. \cite{mohammad2017stance} used feature-based SVM, \cite{wei2016pkudblab} used keyword rules, LitisMind relied on hashtag rules on external data, \cite{dias2016inf} utilized a combination of sentiment classifiers and rules, whereas \cite{krejzl2016uwb} used a maximum entropy classifier with domain-specific features. Our system comfortably surpasses the existing best system at SemEval. Our system manages to improve the existing best system of SemEval 2016 task 6 by 3.2 F-score points for sentiment analysis. 

We also compare our system with the state-of-the-art systems proposed by \cite{schuff2017annotation} on the emotion dataset. The comparison is demonstrated in Table \ref{result-emotion-test}. Maximum entropy, SVM, LSTM, Bi-LSTM, and CNN were the five individual systems used by \cite{schuff2017annotation}. Overall, our proposed system achieves an improvement of 5 F-Score points over the existing state-of-the-art system for emotion analysis. Individually, the proposed system improves the existing F-scores for all the emotions except \textit{surprise}. The findings of \cite{schuff2017annotation} also support this behavior (i.e. worst result for the \textit{surprise} class). This could be attributed to the data scarcity and a very low agreement between the annotators for the emotion \textit{surprise}. 

Experimental results indicate that the multi-task system which uses fine-grained information of emotion analysis helps to boost the performance of 
sentiment analysis. The system M1 comprises of the system S1 performing the main task (sentiment analysis) with E1 undertaking the auxiliary task (emotion analysis). Similarly, the system M2 is made up of S2 and E2 where S2 performs the main task (sentiment analysis) and E2 commits to the auxiliary task (emotion analysis). We observe that in both the situations, the auxiliary task, i.e. emotional information increases the performance of the main task, i.e. sentiment analysis when these two are jointly performed. Experimental results help us to establish the fact that emotion analysis benefits sentiment analysis. The implicit sentiment attached to the emotion words assists the multi-task system. Emotion such as \textit{joy} and \textit{trust} are inherently associated with a \textit{positive} sentiment whereas, \textit{anger}, \textit{disgust}, \textit{fear} and \textit{sadness} bear a \textit{negative} sentiment. Figure \ref{sentiment-fig} illustrates the performance of various models for sentiment analysis.

\begin{table*}[h]
\begin{center}
\resizebox{0.9\textwidth}{!}{
\begin{tabular}{|l|c|c|c|c|c|c|c|c|c|c|}
\hline 
\bf \multirow{2}{*}{Models} & \multirow{2}{*}{Metric} & \multicolumn{9}{|c|}{\textbf{Emotion}} \\ \cline{3-11}
& & Anger & Anticipation & Disgust & Fear & Joy & Sadness & Surprise & Trust & Micro-Avg \\ \hline \hline
\multirow{3}{*}{MaxEnt} & P & 76 & 72 & 62 & 57 & 55 & 65 & 62 & 62 & 66 \\ \cline{2-11}
& R  & 72 & 61 & 47 & 31 & 50 & 65 & 15 & 38 & 52 \\ \cline{2-11}
& F  & 74 & 66 & 54 & 40 & 52 & 65 & 24 & 47 & 58 \\ \hline \hline
\multirow{3}{*}{SVM} & P & 76 & 70 & 59 & 55 & 52 & 64 & 46 & 57 & 63 \\ \cline{2-11}
& R  & 69 & 60 & 53 & 40 & 52 & 60 & 22 & 45 & 53 \\ \cline{2-11}
& F  & 72 & 64 & 56 & 46 & 52 & 62 & 30 & 50 & 58 \\ \hline \hline
\multirow{3}{*}{LSTM} & P & 76 & 68 & 64 & 51 & 56 & 60 & 40 & 57 & 62 \\ \cline{2-11}
& R  & 77 & 68 & 68 & 48 & 41 & 77 & 17 & 49 & 60 \\ \cline{2-11}
& F  & 76 & 67 & 65 & 49 & 46 & 67 & 21 & 51 & 61 \\ \hline \hline
\multirow{3}{*}{BiLSTM} & P & 77 & 70 & 61 & 58 & 54 & 62 & 42 & 59 & 64\\ \cline{2-11}
& R  & 77 & 66 & 64 & 43 & 59 & 72 & 20 & 44 & 60\\ \cline{2-11}
& F  & 77 & \bf 68 & 63 & 49 & 56 & 67 & 27 & 50 & 62\\  \hline \hline
\multirow{3}{*}{CNN} & P & 77 & 68 & 62 & 53 & 54 & 63 & 36 & 53 & 62\\ \cline{2-11}
& R  & 77 & 60 & 61 & 46 & 56 & 72 & 24 & 49 & 59\\ \cline{2-11}
& F  & 77 & 64 & 62 & 49 & 55 & 67 & \bf 28 & 50 & 60\\  \hline \hline
\multirow{3}{*}{\shortstack{E2 \\ (proposed)}} & P & 81 & 74 & 70 & 66 & 64 & 67 & 68 & 68 & 71 \\ \cline{2-11}
& R  & 83 & 62 & 74 & 42 & 59 & 81 & 13 & 49 & 63 \\ \cline{2-11}
& F  & \bf 82 & \bf 68 & \bf 72 & \bf 51 & \bf 62 & \bf 73 & 22 & \bf 57 & \bf 67 \\  \hline
\end{tabular}}
\end{center}
\caption{Comparison with the state-of-the-art systems proposed by \cite{schuff2017annotation} on emotion dataset. The metrics P, R and F stand for Precision, Recall and F1-Score.}
\label{result-emotion-test}
\end{table*}

As a concrete example which justifies the utility of emotion analysis in sentiment analysis is shown below.

\textit{@realMessi he is a real sportsman and deserves to be the skipper.}

The gold labels for the example are \textit{anticipation, joy} and \textit{trust} emotion with a \textit{positive} sentiment. Our system S2 (single task system for sentiment analysis with primary and secondary attention) had incorrectly labeled this example with a negative sentiment and the E2 system (single task system with both primary and secondary attention for emotion analysis) had tagged it with \textit{anticipation} and \textit{joy} only. However, M2 \textit{i.e.} the multi-task system for joint sentiment and emotion analysis had correctly classified the sentiment as \textit{positive} and assigned all the correct emotion tags. It predicted the \textit{trust} emotion tag, in addition to \textit{anticipation} and \textit{joy} (which were predicted earlier by E2). This helped M2 to correctly identify the \textit{positive} sentiment of the example. 
The presence of emotional information helped the system to alter its sentiment decision (\textit{negative} by S2) as it had better understanding of the text. 

A sentiment directly does not invoke a particular emotion always and a sentiment can be associated with more than one emotion. However, emotions like \textit{joy} and \textit{trust} are associated with \textit{positive} sentiment mostly whereas, \textit{anger}, \textit{disgust} and \textit{sadness} are associated with \textit{negative} sentiment particularly. This might be the reason of the extra sentiment information not helping the multi-task system for emotion analysis and hence, a decreased performance for emotion analysis in the multi-task setting.

\begin{table*}[h]
\begin{center}
\resizebox{0.9999\textwidth}{!}{
\parbox{.3\linewidth}{
\begin{center}
\begin{tabular}{|l|c|c|}
\hline 
\multirow{2}{*}{Actual} & \multicolumn{2}{|c|}{Predicted}  \\ \cline{2-3}
 & negative & positive \\ \hline
negative & 1184 & 88 \\ \hline
positive & 236 & 325 \\ \hline
\end{tabular}
\caption{Confusion matrix for \textit{sentiment analysis}}
\label{sentiment}
\end{center}}
\parbox{.45\linewidth}{
\begin{center}
\begin{tabular}{|l|c|c|}
\hline 
\multirow{2}{*}{Actual} & \multicolumn{2}{|c|}{Predicted}  \\ \cline{2-3}
 & NO & YES \\ \hline
NO & 388 & 242 \\ \hline
YES & 201 & 1002 \\ \hline
\end{tabular}
\caption{Confusion matrix for \textit{anger}}
\label{anger}
\end{center}}
\parbox{.45\linewidth}{
\begin{center}
\begin{tabular}{|l|c|c|}
\hline 
\multirow{2}{*}{Actual} & \multicolumn{2}{|c|}{Predicted}  \\ \cline{2-3}
 & NO & YES \\ \hline
NO & 445 & 249 \\ \hline
YES & 433 & 706 \\ \hline
\end{tabular}
\caption{Confusion matrix for \textit{anticipation}}
\label{anticipation}
\end{center}}}

\resizebox{0.9999\textwidth}{!}{
\parbox{.3\linewidth}{
\begin{center}
\begin{tabular}{|l|c|c|}
\hline 
\multirow{2}{*}{Actual} & \multicolumn{2}{|c|}{Predicted}  \\ \cline{2-3}
 & NO & YES \\ \hline
NO & 665 & 277 \\ \hline
YES & 235 & 656 \\ \hline
\end{tabular}
\caption{Confusion matrix for \textit{disgust}}
\label{disgust}
\end{center}}
\parbox{.45\linewidth}{
\begin{center}
\begin{tabular}{|l|c|c|}
\hline 
\multirow{2}{*}{Actual} & \multicolumn{2}{|c|}{Predicted}  \\ \cline{2-3}
 & NO & YES \\ \hline
NO & 911 & 160 \\ \hline
YES & 445 & 317 \\ \hline
\end{tabular}
\caption{Confusion matrix for \textit{fear}}
\label{fear}
\end{center}}
\parbox{.45\linewidth}{
\begin{center}
\begin{tabular}{|l|c|c|}
\hline 
\multirow{2}{*}{Actual} & \multicolumn{2}{|c|}{Predicted}  \\ \cline{2-3}
 & NO & YES \\ \hline
NO & 886 & 236 \\ \hline
YES & 291 & 420 \\ \hline
\end{tabular}
\caption{Confusion matrix for \textit{joy}}
\label{joy}
\end{center}}}
\resizebox{0.9999\textwidth}{!}{
\parbox{.3\linewidth}{
\begin{center}
\begin{tabular}{|l|c|c|}
\hline 
\multirow{2}{*}{Actual} & \multicolumn{2}{|c|}{Predicted}  \\ \cline{2-3}
 & NO & YES \\ \hline
NO & 413 & 405 \\ \hline
YES & 191 & 824 \\ \hline
\end{tabular}
\caption{Confusion matrix for \textit{sadness}}
\label{sadness}
\end{center}}
\parbox{.45\linewidth}{
\begin{center}
\begin{tabular}{|l|c|c|}
\hline 
\multirow{2}{*}{Actual} & \multicolumn{2}{|c|}{Predicted}  \\ \cline{2-3}
 & NO & YES \\ \hline
NO & 1312 & 30 \\ \hline
YES & 426 & 65 \\ \hline
\end{tabular}
\caption{Confusion matrix for \textit{surprise}}
\label{surprise}
\end{center}}
\parbox{.45\linewidth}{
\begin{center}
\begin{tabular}{|l|c|c|}
\hline 
\multirow{2}{*}{Actual} & \multicolumn{2}{|c|}{Predicted}  \\ \cline{2-3}
 & NO & YES \\ \hline
NO & 1032 & 150 \\ \hline
YES & 335 & 316 \\ \hline
\end{tabular}
\caption{Confusion matrix for \textit{trust}}
\label{trust}
\end{center}}}
\end{center}
\end{table*}

\subsection{Error Analysis}
We perform quantitative error analysis for both sentiment and emotion for the M2 model. Table \ref{sentiment} shows the confusion matrix for sentiment analysis. \Cref{anger,anticipation,fear,disgust,joy,sadness,surprise,trust} consist of the confusion matrices for anger, anticipation, fear, disgust, joy, sadness, surprise and trust. We observe from Table \ref{surprise} that the system fails to label many instances with the emotion \textit{surprise}. This may be due to the reason that this particular class is the most underrepresented in the training set. 
A similar trend can also be observed for the emotion \textit{fear} and \textit{trust} in Table \ref{fear} and Table \ref{trust}, respectively. These three emotions have the least share of training instances, making 
the system less confident towards these emotions. 

Moreover, we closely analyze the outputs to understand the kind of errors that our proposed model faces. We observe that the system faces difficulties at times and wrongly predicts the sentiment class in the following scenarios: 

\noindent $\bullet$ Often real-world phrases/sentences have emotions of conflicting nature. These conflicting nature of emotions are directly not evident from the surface form and are left unsaid as these are implicitly understood by humans. The system gets confused when presented with such instances.

\noindent \textbf{Text:} When you become a father you realize that you are not the most important person in the room anymore... Your child is!

\noindent \textbf{Actual Sentiment:} positive \\
\noindent \textbf{Actual Emotion:} anticipation, joy, surprise, trust \\
\noindent \textbf{Predicted Sentiment:} negative \\
\noindent \textbf{Predicted Emotion:} anger, anticipation, sadness

The realization of not being the most important person in a room invokes \textit{anger, anticipation} and \textit{sadness} emotions, and a \textit{negative} sentiment. However, it is a natural feeling of overwhelmingly \textit{positive} sentiment when you understand that your own child is the most significant part of your life.

\noindent $\bullet$ Occasionally, the system focuses on the less significant part of the sentences. Due to this the system might miss crucial information which can influence and even change the final sentiment or emotion. This sometimes lead to the incorrect prediction of the overall sentiment and emotion. 

\noindent \textbf{Text:} I've been called many things, quitter is not one of them...

\noindent \textbf{Actual Sentiment:} positive \\
\noindent \textbf{Actual Emotion:} anticipation, joy, trust \\
\noindent \textbf{Predicted Sentiment:} negative \\
\noindent \textbf{Predicted Emotion:} anticipation, sadness

Here, the system focuses on the first part of the sentence where the speaker was called many things which denotes a \textit{negative} sentiment. Hence, the system predicts a \textit{negative} sentiment and, \textit{anticipation} and \textit{sadness} emotions. However, the speaker in the second part uplifts the overall tone by justifying that s/he has never been called a quitter. This changes the \textit{negative} sentiment to a \textit{positive} sentiment and the overall emotion.

\section{Conclusion}

\noindent In this paper, we have presented a novel two-layered multi-task attention based neural network which performs sentiment analysis through emotion analysis. The primary attention mechanism of the two-layered multi-task system relies on Distributional Thesaurus which acts as a source of external knowledge. The system hierarchically builds the final representation from the word level to the sentence level. This provides a working insight to the system and its ability to handle the unseen words. Evaluation on the benchmark dataset suggests an improvement of 3.2 F-score point for sentiment analysis and an overall performance boost of 5 F-score points for emotion analysis over the existing state-of-the-art systems. The system empirically establishes the fact that emotion analysis is both useful and relevant to sentiment analysis. The proposed system does not rely on any language dependent features or lexicons. This makes it extensible to other languages as well. In future, we would like to extend the two-layered multi-task attention based neural network to other languages.

\section{Acknowledgements}

\noindent Asif Ekbal acknowledges the Young Faculty Research Fellowship (YFRF), supported by Visvesvaraya PhD scheme for Electronics and IT, Ministry of Electronics and Information Technology (MeitY), Government of India, being implemented by Digital India Corporation (formerly Media Lab Asia). 

\bibliographystyle{IEEEtran}
\bibliography{sample-bibliography}

\end{document}